\pdfoutput=1
\documentclass{llncs}
\usepackage[dvipsnames]{xcolor}
\usepackage{capt-of}
\usepackage{algorithm}
\usepackage[noend]{algpseudocode}
\usepackage{environ}
\usepackage{url}
\usepackage{tabularx}
\usepackage{amsmath,amssymb}
\usepackage{tikz}
\usetikzlibrary{matrix,decorations.pathreplacing, calc, positioning}
\usepackage{booktabs,multirow}
\usepackage{filecontents}
\usepackage{nicefrac}
\usepackage{pgfplotstable}
\usepackage{pgfplots}
\usepgfplotslibrary{groupplots}
\usepackage{color, colortbl}
\pgfplotsset{compat=1.14}
\usepackage{caption}
\DeclareMathOperator{\prox}{prox}

\DeclareMathOperator*{\argmin}{arg\,min}

\DeclareMathOperator{\tr}{tr}
\DeclareMathOperator{\pre}{pre}
\DeclareMathOperator{\rec}{rec}

\definecolor{TUGreen}{RGB}{132,184,24}
\definecolor{TUGray}{RGB}{104,104,104}
\definecolor{mygreen}{RGB}{181,221,109}
\definecolor{myyellow}{RGB}{255,237,111}
\definecolor{myseablue}{RGB}{147,213,198}
\definecolor{mylila}{RGB}{189,132,190}
\definecolor{myorange}{RGB}{246,179,98}
\definecolor{myred}{RGB}{247,129,116}
\definecolor{myblue}{RGB}{130,176,208}
\definecolor{myminthe}{RGB}{208,235,189}
\definecolor{mypink}{RGB}{251,205,227}
\definecolor{cPunk}{RGB}{148,103,189}
\definecolor{cPrimp}{RGB}{227,119,194}
\definecolor{cDBSSL1}{RGB}{44,160,44}
\definecolor{cDBSSL2}{RGB}{31,119,180}
\definecolor{cMDLDBSSL}{RGB}{255,127,14}
\newcommand{\R}{\mathbb{R}}

\newcommand{\Ya}{Y^{(a)}}
\newcommand{\Va}{V^{(a)}}
\newcommand{\Da}{D^{(a)}}
\algnewcommand{\IIf}[1]{\State\algorithmicif\ #1\ \algorithmicthen}
\algnewcommand{\EndIIf}{\unskip\ \algorithmicend\ \algorithmicif}
\makeatletter
\newcommand{\problemtitle}[1]{\gdef\@problemtitle{#1}}
\newcommand{\probleminput}[1]{\gdef\@probleminput{#1}}
\newcommand{\problemquestion}[1]{\gdef\@problemquestion{#1}}
\NewEnviron{learningProblem}{
  \problemtitle{}\probleminput{}\problemquestion{}
  \BODY
  \par\addvspace{.5\baselineskip}
  \noindent
  \@problemtitle
  \vspace{-1ex}
  \begin{description}
    \item[Given] \@probleminput 
    \item[Find] \@problemquestion
  \end{description}
  \par\addvspace{.5\baselineskip}
}
\makeatother
\begin{document}

\title{C-SALT: Mining Class-Specific ALTerations in Boolean Matrix Factorization}
%
%
\author{Sibylle Hess \and Katharina Morik}
\authorrunning{Sibylle Hess and Katharina Morik} 
\institute{TU Dortmund University, Computer Science 8, Dortmund, Germany\\
\email{\{sibylle.hess,katharina.morik\}@tu-dortmund.de},\\
\texttt{http://www-ai.cs.uni-dortmund.de/PERSONAL/\{morik,hess\}.html}
}

\maketitle              

\begin{abstract}
Given labeled data represented by a binary matrix, we consider the task to derive a Boolean matrix factorization which identifies commonalities and specifications among the classes. While existing works focus on rank-one factorizations which are either specific or common to the classes, we derive class-specific alterations from common factorizations as well. Therewith, we broaden the applicability of our new method to datasets whose class-dependencies have a more complex structure. On the basis of synthetic and real-world datasets, we show on the one hand that our method is  able to filter structure which corresponds to our model assumption, and on the other hand that our model assumption is justified in real-world application.
Our method is parameter-free.

\keywords{Boolean matrix factorization, shared subspace learning, nonconvex optimization, proximal alternating linearized optimization}
\end{abstract}
\section{Introduction}
When given labeled data, a natural instinct for a data miner is to build a discriminative model that predicts the correct class.
Yet in this paper we put the focus on the characterization of the data with respect to the label, i.e., finding similarities and differences between chunks of data belonging to miscellaneous classes.
Consider a binary matrix where each row is assigned to one class. Such data emerge from fields such as gene expression analysis, e.g., a row reflects the genetic information of a cell, assigned to one tissue type (primary/relapse/no tumor), market basket analysis, e.g., a row indicates purchased items at the assigned store, or from text analyses, e.g., a row corresponds to a document/article and the class denotes the publishing platform. For various applications a characterization of the data with respect to classes is of particular interest. In genetics, filtering the genes which are responsible for the reoccurrence of a tumor may introduce new possibilities for personalized medicine~\cite{sangNature}. In market basket analysis it might be of interest which items sell better in some shops  than others and in text analysis one might ask about variations in the vocabulary used when reporting from diverse viewpoints.
 \begin{figure}
 \centering
 {\tiny
$
\begin{tikzpicture}[decoration=brace,baseline=-0.5ex]
   \matrix [matrix of math nodes,left delimiter=(,right delimiter=)] (n) {
1&1&1&\textcolor{red}{1}&1&1&0&0&0\\
0&0&1&0&1&1&0&0&0\\
1&1&0&\textcolor{red}{1}&0&1&0&0&0\\
1&1&1&\textcolor{red}{1}&1&1&0&0&0\\
0&0&0&0&0&0&1&1&0\\
1&1&0&0&0&1&1&1&\textcolor{red}{1}\\
1&1&0&0&0&1&0&0&\textcolor{red}{1}\\
0&0&0&0&0&0&1&1&0\\
};
\draw[decorate,transform canvas={xshift=-1.5em}] (n-4-1.south west) -- node[left=2pt] {$A$} (n-1-1.north west);
\draw[decorate,transform canvas={xshift=-1.5em}] (n-8-1.south west) -- node[left=2pt] {$B$} (n-5-1.north west);
\draw[color=myblue,fill=myblue,opacity=0.3] (n-1-5.north west) -- (n-1-6.north east) -- (n-2-6.south east) -- (n-2-5.south west) -- (n-1-5.north west);
\draw[color=myblue,fill=myblue,opacity=0.3] (n-4-5.north west) -- (n-4-6.north east) -- (n-4-6.south east) -- (n-4-5.south west) -- (n-4-5.north west);
\draw[color=myblue,fill=myblue,opacity=0.3] (n-1-3.north west) -- (n-1-3.north east) -- (n-2-3.south east) -- (n-2-3.south west) -- (n-1-3.north west);
\draw[color=myblue,fill=myblue,opacity=0.3] (n-4-3.north west) -- (n-4-3.north east) -- (n-4-3.south east) -- (n-4-3.south west) -- (n-4-3.north west);
\draw[color=mygreen,fill=mygreen,opacity=0.3] (n-5-7.north west) -- (n-5-8.north east) -- (n-6-8.south east) -- (n-6-7.south west) -- (n-5-7.north west);
\draw[color=mygreen,fill=mygreen,opacity=0.3] (n-8-7.north west) -- (n-8-8.north east) -- (n-8-8.south east) -- (n-8-7.south west) -- (n-8-7.north west);
\draw[color=myblue,fill=myblue,opacity=0.3] (n-1-3.north west) -- (n-1-3.north east) -- (n-2-3.south east) -- (n-2-3.south west) -- (n-1-3.north west);
\draw[color=myblue,fill=myblue,opacity=0.3] (n-4-3.north west) -- (n-4-3.north east) -- (n-4-3.south east) -- (n-4-3.south west) -- (n-4-3.north west);
\draw[color=mypink,fill=mypink,opacity=0.3] (n-1-1.north west) -- (n-1-2.north east) -- (n-1-2.south east) -- (n-1-1.south west) --(n-1-1.north west);
\draw[color=mypink,fill=mypink,opacity=0.3] (n-1-6.north west) -- (n-1-6.north east) -- (n-1-6.south east) -- (n-1-6.south west) --(n-1-6.north west);
\draw[color=mypink,fill=mypink,opacity=0.3] (n-3-1.north west) -- (n-3-2.north east) -- (n-4-2.south east) -- (n-4-1.south west) --(n-3-1.north west);
\draw[color=mypink,fill=mypink,opacity=0.3] (n-3-6.north west) -- (n-3-6.north east) -- (n-4-6.south east) -- (n-4-6.south west) --(n-3-6.north west);
\draw[color=mypink,fill=mypink,opacity=0.3] (n-6-1.north west) -- (n-6-2.north east) -- (n-7-2.south east) -- (n-7-1.south west) --(n-6-1.north west);
\draw[color=mypink,fill=mypink,opacity=0.3] (n-6-6.north west) -- (n-6-6.north east) -- (n-7-6.south east) -- (n-7-6.south west) --(n-6-6.north west);
\end{tikzpicture}
\approx
\begin{tikzpicture}[baseline=-0.5ex]
    \matrix [matrix of math nodes,left delimiter=(,right delimiter=)] (n) {
1&1&0 \\
0&1&0 \\
1&0&0 \\
1&1&0 \\
0&0&1 \\
1&0&1 \\
1&0&0\\
0&0&1 \\
};
\draw[color=myblue,fill=myblue,opacity=0.3] (n-1-2.north west) -- (n-1-2.north east) -- (n-2-2.south east) -- (n-2-2.south west) -- (n-1-2.north west);
\draw[color=myblue,fill=myblue,opacity=0.3] (n-4-2.north west) -- (n-4-2.north east) -- (n-4-2.south east) -- (n-4-2.south west) -- (n-4-2.north west);
\draw[color=mygreen,fill=mygreen,opacity=0.3] (n-5-3.north west) -- (n-5-3.north east) -- (n-6-3.south east) -- (n-6-3.south west) -- (n-5-3.north west);
\draw[color=mygreen,fill=mygreen,opacity=0.3] (n-8-3.north west) -- (n-8-3.north east) -- (n-8-3.south east) -- (n-8-3.south west) -- (n-8-3.north west);
\draw[color=mypink,fill=mypink,opacity=0.3] (n-1-1.north west) -- (n-1-1.north east) -- (n-1-1.south east) -- (n-1-1.south west) --(n-1-1.north west);
\draw[color=mypink,fill=mypink,opacity=0.3] (n-3-1.north west) -- (n-3-1.north east) -- (n-4-1.south east) -- (n-4-1.south west) --(n-3-1.north west);
\draw[color=mypink,fill=mypink,opacity=0.3] (n-6-1.north west) -- (n-6-1.north east) -- (n-7-1.south east) -- (n-7-1.south west) --(n-6-1.north west);
\end{tikzpicture}
\cdot
\begin{tikzpicture}[baseline=-0.5ex]
    \matrix [matrix of math nodes,left delimiter=(,right delimiter=)] (n) {
1&1&0&0&0&1&0&0&0 \\
0&0&1&0&1&1&0&0&0 \\
0&0&0&0&0&0&1&1&0 \\
};
\draw[color=myblue,fill=myblue,opacity=0.3] (n-2-3.north west) -- (n-2-3.north east) -- (n-2-3.south east) -- (n-2-3.south west) -- (n-2-3.north west);
\draw[color=myblue,fill=myblue,opacity=0.3] (n-2-5.north west) -- (n-2-6.north east) -- (n-2-6.south east) -- (n-2-5.south west) -- (n-2-5.north west);
\draw[color=mygreen,fill=mygreen,opacity=0.3] (n-3-7.north west) -- (n-3-8.north east) -- (n-3-8.south east) -- (n-3-7.south west) -- (n-3-7.north west);
\draw[color=mypink,fill=mypink,opacity=0.3] (n-1-1.north west) -- (n-1-2.north east) -- (n-1-2.south east) -- (n-1-1.south west) --(n-1-1.north west);
\draw[color=mypink,fill=mypink,opacity=0.3] (n-1-6.north west) -- (n-1-6.north east) -- (n-1-6.south east) -- (n-1-6.south west) --(n-1-6.north west);
\end{tikzpicture}
$
}
 \caption{A Boolean factorization of rank three. The data matrix on the left is composed by transactions belonging to two classes $A$ and $B$. Each outer product is highlighted. Best viewed in color.}
 \label{fig:classFact}
 \end{figure}

These questions are approached as pattern mining~\cite{characterDifference} and Boolean matrix factorization problems~\cite{MiettinenJSBMF}. Both approaches search for factors or patterns which occur in both or only one of the classes. This is illustrated in Fig.~\ref{fig:classFact}; a data matrix is indicated on the left, whose rows are assigned to one class, $A$ or $B$. While the pink outer product spreads over both classes, the blue and green products concentrate in only one of the classes. We refer to the factorizations of the first kind as common and to those of the second kind as class-specific.

The identification of class specific and common factorizations is key to a characterization of similarities and differences among the classes. Yet, what if meaningful deviations between the classes are slightly hidden underneath an overarching structure? The factorization in Fig.~\ref{fig:classFact} is not exact, we can see that the red colored ones in the data matrix are not taken into account by the model. This is partially desired as the data is expected to contain noise which is supposedly filtered by the model. On the other hand, we can observe concurrence of the red ones and the pink factors -- in each class.
\subsection{Main Contributions}
In this paper we propose a novel Boolean Matrix Factorization (BMF) method which is suitable to compare horizontally concatenated binary data matrices originating from diverse sources or belonging to various classes. To the best of the authors knowledge, this is the first method in the field of matrix factorizations of any kind, combining the properties listed below in one framework:
\begin{enumerate}
\item \label{contrib:classes}the method can be applied to compare any number of classes or sources,
\item \label{contrib:rank}the factorization rank is automatically determined; this includes the number of outer products, which are common among multiple classes, but also the number of discriminative outer products occurring in only one class,
\item \label{contrib:V}in addition to discriminative rank-one factorizations, more subtle characteristics of classes can be derived, pointing out how common outer products deviate among the classes.
\end{enumerate}
While works exist which approach one of the points~\ref{contrib:classes} or~\ref{contrib:rank} (see Sec.~\ref{sec:relatedWork}), the focus on subtle deviations among the classes as addressed in point~\ref{contrib:V} is entirely new. This expands the applicability of the new method to datasets where deviations among the classes have a more complex structure.
\section{Preliminaries}
We identify items $\mathcal{I}=\{1,\ldots,n\}$ and transactions $\mathcal{T}=\{1,\ldots,m\}$ by a set of indices of a binary matrix $D\in \{0,1\}^{m\times n}$. This matrix represents the data, having  $D_{ji}=1$ iff transaction $j$ contains item $i$. A set of items is called a \emph{pattern}.

We assume that the data matrix is composed of various sources, identified by an assignment from transactions to classes. Denoting by $[A^{(a)}]_a$ the matrix vertically concatenating the matrices $A^{(a)}$ for $a\in\{1,\ldots,c\}$, we write
\begin{align}\label{eq:DaYaVa}
D=\begin{bmatrix}
D^{(a)}
\end{bmatrix}_a,\
Y=\begin{bmatrix}
Y^{(a)}
\end{bmatrix}_a \text{ and }
V^T=\begin{bmatrix}
V^{(a)T}
\end{bmatrix}_a.
\end{align}
The ($m_a\times n$)-matrix $D^{(a)}$ comprises the $m_a<m$ transactions belonging to class $a$. Likewise, we explicitly notate the class-related $(m_a\times r)$- and $(n\times r)$-dimensional parts of the $m\times r$ and $n\times rc$ factor matrices $Y$ and $V$ as $Y^{(a)}$ and $\Va$. These factor matrices are properly introduced in Sec.~\ref{sec:problemdef}. 

We often employ the function $\theta_t$ which rounds a real value $x\geq t$ to one and $x< t$ to zero. We abbreviate $\theta_{0.5}$ to $\theta$ and denote with $\theta(X)$ the entry-wise application of $\theta$ to a matrix $X$.
We denote matrix norms as $\|\cdot\|$ for the Frobenius norm and $|\cdot|$ for the entry-wise 1-norm. 
We express with $x^{m\times n}$ the $(m\times n)$-dimensional matrix having all entries equal to $x$. The operator $\circ$ denotes the Hadamard product. Finally, we denote with $\log$ the natural logarithm.
\subsection{Boolean Matrix Factorization in Brief}\label{sec:bmf}
Boolean Matrix Factorization (BMF) assumes that the data $D\in\{0,1\}^{m\times n}$ originates from a matrix product with some noise, i.e.,
\begin{align}\label{eq:BMF}
D=\theta(YX^T)+N,
\end{align}
where $X\in\{0,1\}^{n\times r}$ and $Y\in\{0,1\}^{m\times r}$ are the factor matrices of rank $r$ and $N\in\{-1,0,1\}^{m\times n}$ is the noise matrix. The Boolean product conjuncts $r$ matrices; the outer products $Y_{\cdot s}X_{\cdot s}^T$ for $1\leq s\leq r$. We use $\theta$ to denote the Boolean conjunction in terms of elementary algebra. Each outer product is defined by a pattern, indicated by $X_{\cdot s}$, and a set of transactions using the pattern, indicated by $Y_{\cdot s}$. Correspondingly, $X$ is called the pattern and $Y$ the usage matrix.

Unfortunately, solving $X$ and $Y$ from Eq.~(\ref{eq:BMF}), if only the data matrix $D$ is known, is generally not possible. Hence, surrogate tasks are formulated in which the data is approximated by a matrix product according to specific criteria. The most basic approach is to find the factorization of given rank which minimizes the residual sum of absolute values $|D-\theta(YX^T)|$. This problem, however, cannot be approximated within any factor in polynomial time (unless $\mathbf{NP}= \mathbf{P}$)~\cite{discreteBasisProb}.

BMF has a very popular relative, called  Nonnegative Matrix Factorization (NMF). Here, a nonnegative data matrix $D\in\R_+^{m\times n}$ is approximated by the product of nonnegative matrices $X\in\R_+^{n\times r}$ and $Y\in\R_+^{m\times r}$. NMF tasks often involve minimizing the Residual Sum of Squares (RSS)  $\frac{1}{2}\|D-YX^T\|^2$~\cite{wang2013nmfReview}. Minimizing the RSS subject to binary matrices $X$ and $Y$ introduces the task of binary matrix factorization~\cite{zhangApplication}.
\begin{figure}[!t]
\centering
{\tiny
$
\begin{tikzpicture}[decoration=brace,baseline=-0.5ex]
   \matrix [matrix of math nodes,left delimiter=(,right delimiter=), every node/.append style={text width=0.2cm,align=center,minimum height=5ex},
  nodes in empty cells,] (n) {
&&&&\\
&&&&\\
&&&&\\
&&&&\\
&&&&\\
&&&&\\
};
\node[font=\normalsize] 
  at ([xshift=-1pt, yshift=-3pt]n-3-1) {$Y_{S}$};
\node[font=\normalsize] 
  at ([xshift=-5pt, yshift=-2pt]n-2-3) {$Y^{(1)}_{D}$};
  \node[font=\normalsize] 
  at ([xshift=-5pt, yshift=-2pt]n-5-5) {$Y^{(2)}_{D}$};
\draw[decorate,transform canvas={xshift=-1.5em}] (n-3-1.south west) -- node[left=2pt] {$m_1$} (n-1-1.north west);
\draw[decorate,transform canvas={xshift=-1.5em}] (n-6-1.south west) -- node[left=2pt] {$m_2$} (n-4-1.north west);
\draw[decorate,transform canvas={yshift=-1em}] (n-6-1.south east) -- node[below=2pt] {$r_0$} (n-6-1.south west);
\draw[decorate,transform canvas={yshift=-1em}] (n-6-3.south east) -- node[below=2pt] {$r_1$} (n-6-2.south west);
\draw[decorate,transform canvas={yshift=-1em}] (n-6-5.south east) -- node[below=2pt] {$r_2$} (n-6-4.south west);
\draw[color=myblue,fill=myblue,opacity=0.3] (n-1-2.north west) -- (n-1-3.north east) -- (n-3-3.south east) -- (n-3-2.south west) -- (n-1-2.north west);
\draw[color=mygreen,fill=mygreen,opacity=0.3] (n-4-4.north west) -- (n-4-5.north east) -- (n-6-5.south east) -- (n-6-4.south west) -- (n-4-4.north west);
\draw[color=mypink,fill=mypink,opacity=0.3] (n-1-1.north west) -- (n-1-1.north east) -- (n-6-1.south east) -- (n-6-1.south west) --(n-1-1.north west);
\end{tikzpicture}
\qquad
\begin{tikzpicture}[decoration=brace,baseline=-0.5ex]
   \matrix [matrix of math nodes,left delimiter=(,right delimiter=), every node/.append style={text width=0.2cm,align=center,minimum height=5ex},
  nodes in empty cells,] (n) {
&&&&&&\\
&&&&&&\\
&&&&&&\\
&&&&&&\\
&&&&&&\\
};
\node[font=\normalsize] 
  at (n-1-4) {$X_{S}^T$};
\node[font=\normalsize] 
  at ([yshift=-6pt]n-2-4) {$X_1^T$};
  \node[font=\normalsize] 
  at ([yshift=-6pt]n-4-4) {$X_2^T$};
\draw[decorate,transform canvas={xshift=-1.5em}] (n-1-1.south west) -- node[left=2pt] {$r_0$} (n-1-1.north west);
\draw[decorate,transform canvas={xshift=-1.5em}] (n-3-1.south west) -- node[left=2pt] {$r_1$} (n-2-1.north west);
\draw[decorate,transform canvas={xshift=-1.5em}] (n-5-1.south west) -- node[left=2pt] {$r_2$} (n-4-1.north west);
\draw[decorate,transform canvas={yshift=-1em}] (n-5-7.south east) -- node[below=2pt] {$n$} (n-5-1.south west);
\draw[color=myblue,fill=myblue,opacity=0.3] (n-2-1.north west) -- (n-2-7.north east) -- (n-3-7.south east) -- (n-3-1.south west) -- (n-2-1.north west);
\draw[color=mygreen,fill=mygreen,opacity=0.3] (n-4-1.north west) -- (n-4-7.north east) -- (n-5-7.south east) -- (n-5-1.south west) -- (n-4-1.north west);
\draw[color=mypink,fill=mypink,opacity=0.3] (n-1-1.north west) -- (n-1-7.north east) -- (n-1-7.south east) -- (n-1-1.south west) --(n-1-1.north west);
\end{tikzpicture}
$
}
\caption{A Boolean product identifying common (pink) and class-specific outer products (blue and green). Best viewed in color.}
\label{fig:usageJSMF}
\end{figure}
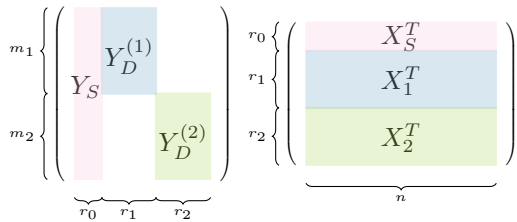
\subsection{Related Work}\label{sec:relatedWork}
If the given data matrix is class-wise concatenated (cf.\@ Eq.~(\ref{eq:DaYaVa})), a first approach for finding class-defining characteristics is to separately derive factorizations for each class. However, simple approximation measurements as discussed in Sec.~\ref{sec:bmf} are already nonconvex and have multiple local optima.
Due to this vagueness of computed models, class-wise factorizations are not easy to interpret; they lack a view on the global structure. Puzzling together the (parts of) patterns defining (dis-)similarities of classes afterwards, is non-trivial.

In the case of nonnegative, labeled data matrices, measures such as Fisher's linear discriminant criterion are minimized to derive weighted feature vectors, i.e., patterns in the binary case, which discriminate most between classes. This variant of NMF is successfully implemented for classification problems such as face recognition~\cite{nikitidis2014discrimNMF} and identification of cancer-associated genes~\cite{odibat2014discrNMFGeneExpr}.

For social media retrieval, Gupta et al. introduce Joint Subspace Matrix Factorization (JSMF)~\cite{guptaJSMF2010}. Focusing on the two-class setting, they assume that data points (rows of the data matrix) emerge not only from discriminative but also from common subspaces. JSMF infers for a given nonnegative data matrix and ranks $r_0,r_1$ and $r_2$ a factorization as displayed in Fig.~\ref{fig:usageJSMF}. Multiplicative updates minimize the weighted sum of class-wise computed RSS.
In Regularized JSNMF (RJSNMF), a regularization term is used to prevent that shared feature vectors swap into discriminative subspaces and vice versa~\cite{gupta2013RJSNMF}. The arising optimization problem is solved by the method of Lagrange multipliers. Furthermore, a provisional method to determine the rank automatically is evaluated. However, this involves multiple runs of the algorithm with increasing rank of shared and discriminative subspaces, until the approximation error barely decreases.
A pioneering extension to the multi-class case  is provided in~\cite{gupta2014matrix}.

Miettinen~\cite{MiettinenJSBMF} transfers the objective of JSMF into Boolean algebra, solving
\[\min_{X,Y}\sum_{a\in\{1,2\}}\frac{\mu_a}{2}\left|\Da-\theta\left(\begin{bmatrix}
\Ya_{S} & \Ya_D
\end{bmatrix} \begin{bmatrix}
	X_S^T\\
    X_a^T
\end{bmatrix}\right)\right|\]
for binary matrices $D, X$ and $Y$, and normalizing constants $\mu_{1/2}^{-1} =|D^{(2/1)}|$.  A variant of the BMF algorithm \textsc{Asso}~\cite{discreteBasisProb} governs the minimization. A provisional determination of ranks based on the Minimum Description Length (MDL) principle is proposed, computing which of the candidate rank constellations yields the lowest description length. The description length captures model complexity and data fit, and is hence suitable for model order selection \cite{miettinen2011model,primp}.

Budhatoki and Vreeken~\cite{characterDifference} pursue the idea of MDL to derive a set of pattern sets, which characterizes similarities and differences of groups of classes. Identifying the usage of each pattern with its support in the data, the number of derived patterns equates the rank in BMF. In this respect, their proposed algorithm \textsc{DiffNorm} automatically determines the ranks in the multi-class case. However, the posed constraint on the usage often results in vast amount of returned patterns.    

In the two-class nonnegative input matrices case, Kim et al. improve over RJSNMF by allowing small deviations from shared patterns in each class~\cite{kim2015JNMF}. They found that shared patterns are often marginally altered according to the class. In this paper, we aim at finding these overlooked variations of shared patterns together with strident differences among multiple classes, combining the strengths of MDL for rank detection and the latest results in NMF.
\subsection{(Informal) Problem Definition}\label{sec:problemdef}
Given a binary data matrix composed from multiple classes, we assume that the data has an underlying model similar to the one in Fig.~\ref{fig:classFact}. There are common or shared patterns (pink) and class-specific patterns (blue and green). Furthermore, there are class-specific patterns, which align within a subset of the classes where a pattern is used (the red ones). We call such aligning patterns class-specific alterations and introduce the matrix $V$ to reflect these.
\begin{definition}\label{def:classSpec}
Let $X\in\{0,1\}^{n\times r}$ and $V\in \{0,1\}^{n\times cr}$.
We say the matrix $V$ models \emph{class-specific alterations of $X$} if $\|X\circ \Va\|=0$ for all $1\leq a\leq c$, and $\|V^{(1)}\circ\ldots\circ V^{(c)}\|=0$.
\end{definition}
Similar to the data decomposition denoted in Eq.~(\ref{eq:BMF}), we assume that data emerges from a Boolean matrix product; yet, we now consider multiple products, one for each class, which are defined by the class-wise alteration matrix $V$, its pattern matrix, usage and the noise matrix $N=[N^{(a)}]_a$, such that for $1\leq a\leq c$
\begin{align}\label{eq:VBMF}
\Da= \theta\left(\Ya(X+\Va)^T\right) +N^{(a)}.
\end{align}
Given a class-wise composed binary data matrix, we consider the task to filter the factorization, defined by $X$, $Y$ and $V$, from the noise.
\section{The Proposed Method}
We build upon the BMF algorithm \textsc{Primp}, which combines recent results from numerical optimization with MDL in order to return interpretable factorizations of a suitably estimated rank~\cite{primp}. The employed description length $f$ reflects the size of the data encoded by a code table as known from algorithms \textsc{Slim} and \textsc{Krimp}~\cite{siebes2006item,Slim}. Determining a smooth function $F$, bounding the description length from above, and a function $\phi$ to penalize non-binary values, locally minimizing matrices of the relaxed objective $F(X,Y)+ \phi(X) +\phi(Y)$ are derived. Rounding the local minimizers to binary matrices according to the description length, yields the final result and decides over the rank of the factorization.

The numerical optimization is performed by \emph{Proximal Alternating Linearized Minimization} (PALM)~\cite{teb14}. That are alternatingly invoked \emph{proximal mappings} with respect to $\phi$ from the gradient descent update with respect to $F$ (cf.\@ lines~\ref{alg:proxX},\ref{alg:proxV} and \ref{alg:proxY} in Algorithm~\ref{alg:C-Salt}).
The proximal mapping of $\phi$ returns a matrix satisfying the following minimization criterion:
\[\prox_\phi(X) \in \argmin_{\hat{X}}\left\{\frac{1}{2}\|X-\hat{X}\|^2+\phi(\hat{X})\right\}.\]
Loosely speaking, $X$ is given a little push into a direction minimizing $\phi$.
We choose $\phi(X)=\sum_{i,j}\Lambda(X_{ij})$ to penalize non-binary matrix-entries by an entry-wise application of the function $\Lambda$. Correspondingly, the prox-operator is computed entry-wise $\prox_{\alpha\phi}(X)=(\prox_{\alpha\Lambda}(X_{ji}))_{ji}$, where
\[
	\Lambda(x) =
    \begin{cases}
        -|1-2x|+1 &x\in[0,1]\\
        \infty &x\notin[0,1].
    \end{cases},\
    \prox_{\alpha \Lambda}(x)=
    \begin{cases}
        \max\{0,x-2\alpha\} &x\leq0.5\\
        \min\{1,x+2\alpha\} &x>0.5.
    \end{cases}
\]
 Notice, the proximal mapping ensures that factor matrices always attain values between zero and one. For further information on prox-operators, see, e.g., \cite{parikh2014proximal}.

The step sizes of the gradient descent updates are computed by the Lipschitz moduli of partial gradients (cf.\@ lines ~\ref{alg:stepX}, \ref{alg:stepV} and \ref{alg:stepY} in Algorithm~\ref{alg:C-Salt}).
Assuming that the infimum of $F$ and $\phi$ exists and $\phi$ is proper and lower continuous, PALM generates a nonincreasing sequence of function values which converges to a critical point of the relaxed objective.
\subsection{C-Salt}\label{sec:CSalt}
In order to capture class-defining characteristics in the framework of \textsc{Primp}, few extensions have to be made.
We pose two requirements on the interplay between usage and class-specific alterations of patterns: class-specific alterations ought to fit very well to the corresponding class but as little as possible to other classes.
We introduce a regularizing function to penalize nonconformity to this request.
\begin{align*}
S(Y,V) &= \sum_{s=1}^r\sum_{a=1}^c \left(\left|\Ya_{\cdot s}\right|\left|\Va_{\cdot s}\right|-{\Ya}^T\Da \Va_{\cdot s}\right)+\sum_{b\neq a}{Y^{(b)}_{\cdot s}}^TD^{(b)}\Va_{\cdot s}\\
&=\sum_{a=1}^c\tr\left(\left({\Ya}^T(1^{m_a\times n}-2\Da) + Y^TD\right)\Va\right).
\end{align*}

We extend the description length of \textsc{Primp} such that class-specific alterations are encoded in the same way as patterns; by standard codes, assigning item $i\in\mathcal{I}$ a code of length $u_i=-\log\left(\nicefrac{|D_{\cdot i}|}{|D|}\right)$. The objective function $f$ adds the description length to the specificity-regularizer
\begin{align*}
f(X,V,Y)&=
-\sum_{s:|Y_{\cdot s}|> 0} \left((|Y_{\cdot s}|{+}1) \cdot \log\left(\frac{|Y_{\cdot s}|}{|Y|+|N|}\right) +X_{\cdot s}^Tu+\sum_{a}{\Va_{\cdot s}}^Tu\right)\\
&\quad -\sum_{i:|N_{\cdot i}|> 0} \left((|N_{\cdot i}|{+}1) \cdot \log\left(\frac{|N_{\cdot i}|}{|Y|+|N|}\right) +u_i\right)
+S(Y,V).
\end{align*}
This determines the relaxed objective $F(X,V,Y) +\phi(X) +\phi(V)+\phi(Y)$, where
\begin{align*}
F(X,V,Y)&= \frac{1}{2}\left(\mu\sum_{a=1}^c\|\Da-\Ya(X{+}\Va)^T\|^2
+ G(X,V,Y)
+S(Y,V)\right),
\end{align*}
$\mu=1+\log(n)$ and $G$ is defined as stated in  Appendix~\ref{app:func}.
$F$ has Lipschitz continuous gradients and is suitable for PALM.

\begin{algorithm}[t]
\caption{\textsc{C-Salt}($D=[D^{(a)}]_a;\Delta_r=10,\gamma=1.00001$)$\qquad$}
\begin{algorithmic}[1]
  \State $(X_K,V_K,Y_K)\gets (\emptyset, \emptyset, \emptyset)$
  \For {$r\in\{\Delta_r,2\Delta_r,3\Delta_r,\ldots\}$}
    \State $(X_0,V_0,Y_0) \gets $\Call{IncreaseRank}{$X_K, V_K, Y_K,\Delta_r$} \Comment{Append random columns}
    \For {$k = 0,1,\ldots$}\label{alg:optStart} \Comment{Select stop criterion}
      \State $\nicefrac{1}{\alpha_k} \gets \gamma M_{\nabla_XF}(V_k,Y_k)$ \label{alg:stepX}
      \State $X_{k+1} \gets \prox_{\alpha_k\phi}\left(X_k-\alpha_k\nabla_XF(X_k,V_k,Y_k)\right)$ \label{alg:proxX}
      \State $\nicefrac{1}{\nu_k^{(a)}} \gets \gamma M_{\nabla^{(a)}_VF}(X_{k+1},Y_k)$  \Comment{$1\leq a\leq c$} \label{alg:stepV}
      \State $\Va_{k+1} \gets \prox_{\nu_k^{(a)}\phi}\left(\Va_k-\nu_k^{(a)}\nabla^{(a)}_VF(X_{k+1},V_{k}^{(a)},Y_k)\right)$ \Comment{$1\leq a\leq c$} \label{alg:proxV}
      \State $\nicefrac{1}{\beta_k} \gets\gamma M_{\nabla_YF}(X_{k+1},V_{k+1})$  \label{alg:stepY}
      \State $Y_{k+1} \gets \prox_{\beta_k\phi}\left(Y_k-\beta_k\nabla_YF(X_{k+1},V_{k+1},Y_k)\right)$ \label{alg:proxY}
    \EndFor\label{alg:optEnd}
    \State $(X,V,Y)\gets \Call{Round}{f,X_k,V_k,Y_k}$\Comment{Try thresholds from finite set}
    \IIf {$r-r(X,V,Y)>1$}\label{alg:rankStart}
    	\textbf{return} $(X,V,Y)$
    \EndIIf\label{alg:rankEnd}
    \EndFor
\end{algorithmic}
\label{alg:C-Salt}
\end{algorithm}

Algorithm~\ref{alg:C-Salt} details \textsc{C-Salt}, which largely follows the framework of \textsc{Primp}~\cite{primp}. \textsc{C-Salt} has as input the data $D$ and two parameters, for which default values are given, which rarely need to be adjusted in practice. Further information about the robustness and significance of these parameters is provided in~\ref{alg:C-Salt}. For step-wise increased ranks, PALM optimizes the relaxed objective (lines \ref{alg:optStart}-\ref{alg:optEnd}). Note that the alternating minimization of more than two matrices corresponds to the extension of PALM for multiple blocks, discussed in~\cite{teb14}. The required gradients and Lipschitz moduli are stated in Appendix~\ref{app:func}. Subsequently, a rounding procedure  returns the binary matrices $X_{t_1}=\theta_{t_1}(X_K),\ V_{t_1}=\theta_{t_1}(V_K)$  and $Y_{t_2}=\theta_{t_2}(Y_K)$ for thresholds $t_1,t_2\in\{0.05k\mid k\in\{0,1,\ldots,20\}\}$ minimizing $f$. Thereby, the validity of Definition~\ref{def:classSpec} is ensured by setting unsuitable values in $V$ to zero. Furthermore, \textit{trivial} outer products covering fewer than two transactions or items are removed. The number of remaining outer products defines the rank $r(X,V,Y)$. If the gap between the number of possibly and actually modeled outer products is larger than one, the current factorization is returned (line~\ref{alg:rankStart}). 
\section{Experiments}
The experimental evaluations concern the following research questions:
\begin{enumerate}
\item Given that the data matrix is generated as stated by the informal problem definition in Sec.~\ref{sec:problemdef}, does \textsc{C-Salt} find the original data structure?
\item Is the assumption that real-world data emerge as stated in Eq.~(\ref{eq:VBMF}) reasonable, and what effect has the modeling of class-specific alterations on the results?
\end{enumerate}
We compare against the algorithms \textsc{Dbssl}, the dominated approach proposed in~\cite{MiettinenJSBMF}, and \textsc{Primp}\footnote{\url{http://sfb876.tu-dortmund.de/primp}}.
The first question is approached by a series of synthetic datasets, generated according to Eq.~(\ref{eq:VBMF}). To address the second question, we compare on real-world datasets the RSS, computed factorization ranks and visually inspect derived patterns.
Furthermore, we discuss an application in genome analysis where none of the existing methods provides the crucial information.

For \textsc{C-Salt} and \textsc{Primp} we use as stop criterion a minimum average function decrease (of last 500 iterations) of 0.005 and maximal $k\leq10,000$ iterations. We use the Matlab/C implementation of \textsc{Dbssl} which has been kindly provided by the authors upon request. Setting the minimum support parameter of the employed FP-Growth algorithm proved tricky. Choosing the minimum support too low results in a vast memory consumption (we provided 100GiB RAM); setting it too high yields too few candidate patterns. Hence, this parameter varies between experiments within the range \{2,\ldots,8\}.

\textsc{C-Salt} is implemented for GPU, as is \textsc{Primp}. We provide the source code of our algorithms together with the data generating script \footnote{\url{http://sfb876.tu-dortmund.de/csalt}}.
\subsection{Measuring the Quality of Factorizations}\label{sec:QualityMeas}
For synthetic datasets, we compare the computed models against the planted structure by an adaptation of the micro-averaged F-measure.
We assume that generated matrices  $X^\star,V^\star,Y^\star$ and computed models $X,V,Y$ have the same rank $r$. Otherwise, we attach columns of zeros to make them match.
We compute one-to-one matchings $\sigma_1:\{1,\ldots, r\}\rightarrow \{1,\ldots,r\}$ between outer products of computed and generated matrices by the Hungarian algorithm~\cite{hungarian}. The matching maximizes $\sum_{s=1}^rF^{(a)}_{s,\sigma_1(s)}$, where 
\[
	F^{(a)}_{S,T}=2\frac{\pre^{(a)}_{S,T}\cdot\rec^{(a)}_{S,T}}{\pre^{(a)}_{S,T}+\rec^{(a)}_{S,T}},
\]
for selections of columns $S$ and $T$. $\pre^{(a)}_{S,T}$ and $\rec^{(a)}_{S,T}$ denote precision and recall w.r.t. the denoted column selection. Writing $X^{(a)}=X+V^{(a)}$, we compute
\begin{align*}
	\pre^{(a)}_{S,T} &= \frac{\left|\left({Y^\star_{\cdot S}}\circ Y_{\cdot T}\right)^{(a)}\left({X^\star_{\cdot S}}\circ X_{\cdot T}\right)^{(a)^T}\right|}{\left|\Ya_{\cdot T}{X^{(a)}_{\cdot T}}^T\right|}, &
    \rec^{(a)}_{S,T} &= \frac{\left|\left({Y^\star_{\cdot S}}\circ Y_{\cdot T}\right)^{(a)}\left({X^\star_{\cdot S}}\circ X_{\cdot T}\right)^{(a)^T}\right|}{\left|{Y^\star_{\cdot S}}^{(a)}{{X^\star_{\cdot S}}^{(a)}}^T\right|}.
\end{align*}
We calculate then precision and recall such that planted outer products with indices $R=(1,\ldots, r)$ are compared to outer products of the computed factorization with indices $\sigma_1(R)=(\sigma_1(1),\ldots,\sigma_1(r))$. The corresponding $F$-measure is the micro $F$-measure, which is identified by $F^{(a)}_{R,\sigma_1(R)}$.

Since class-specific alterations of patterns, reflected by the matrix $V$, are particularly interesting in the scope of this paper, we additionally state the recall of $V^\star$, denoted by $\rec_V$. Therefore, we compute a maximum matching $\sigma_2$ between generated class alterations $V^\star$ with usage $Y^\star$ and computed patterns $X_V=[X\ V]$ (setting $V$ to the $(n\times cr)$ zero matrix for other algorithms than \textsc{C-Salt}) with usage $Y_V=[Y \ldots Y]$ (concatenating $c$ times). The recall $\rec_{R,\sigma_2(R)}^{(a)}$ is then computed with respect to the matrices $V^\star, Y^\star, X_V$ and $Y_V$.
Furthermore, we compute the class-wise factorization rank $r^{(a)}$ as the number of nontrivial outer products, involving more than only one column or row. Outer products where solely one item or one transaction is involved yield no insight for the user and are therefore always discarded. In following plots, we indicate averaged measures over all classes
\[ F=\frac{1}{c}\sum_a F^{(a)}_{R,\sigma_1(R)}, \quad \rec_V=\frac{1}{c}\sum_a \rec^{(a)}_{R,\sigma_2(R)} \text{ and } r=\frac{1}{c}\sum_a r^{(a)}.
\]
Therewith, the size of the class is not taken into account; the discovery of planted structure is considered equally important for every class.
$F$-measure and recall have values between zero and one. The closer both approach one, the more similar are the obtained and planted factorizations.
\subsection{Synthetic Data Generation}\label{sec:SynthGen}
We state the synthetic data generation as a procedure which receives the matrix dimensions $(m_a)_a$ ($m=\sum_am_a$) and $n$, the factorization rank $r^\star$, matrix $C\in\{0,1\}^{c\times r}$ and noise probability $p$ as input. The matrix $C$ indicates for each pattern in which classes it is used.
\begin{description}
\item [\textbf{GenerateData}($n,(m_a)_a,r^\star,C,p$)]
\item 1. Draw the $(n\times r^\star)$ and $(m\times r^\star)$ matrices $X^\star$, ${\Va}^\star$ and $Y^\star$ uniformly random from the set of all  binary matrices subject to
\begin{itemize}
	\item each column $X^\star_{\cdot s} (Y^\star_{\cdot s})$ has at least $\nicefrac{n}{100}(\nicefrac{m}{100})$ uniquely assigned bits,
	\item the density is bounded by $|X^\star_{\cdot s}|\leq \nicefrac{n}{10}$ and $|{\Ya_{\cdot s}}^\star|\leq C_{sa}\nicefrac{m_a}{10}$
	\item ${\Va}^\star$ models class-specific alterations of $X^\star$ and
$\left|\sum_{a=1}^c{\Va_{s}}^\star\right|\leq \nicefrac{2}{3} \left|X^\star_{\cdot s}\right|$
\end{itemize}
\item 3. Set $\Da$, flipping every bit of $\theta\left({\Ya}^\star (X^\star+{\Va}^\star)^T\right)$ with probability $p$.
\end{description}
By default, the parameters $r^\star=24$, $m_a=\nicefrac{m}{2}$, where $m$ and $n$ are varied as described in Sec.~\ref{sec:synthExp}, $p=0.1$, and depending on the number of classes we set
\[C_2=\left[
\begin{pmatrix}
1 & 0 & 1\\
1 & 1 & 0\\
\end{pmatrix}\right]_{\frac{r^\star}{3}},\quad C_3=\left[\begin{pmatrix}
1 & 1 & 0 & 0\\
1 & 0 & 1 & 0\\
1& 0 & 1 & 1\\
\end{pmatrix}\right]_{\frac{r^\star}{4}},\quad
C_4=\left[\begin{pmatrix}
1 & 1 & 0 & 0 & 0\\
1 & 0 & 1 & 0 & 0\\
1 & 0 & 1 & 1 & 0\\
1 & 0 & 1 & 1 & 1\\
\end{pmatrix}\right]_{\frac{r^\star}{5}} .\]
\subsection{Synthetic Data Experiments}\label{sec:synthExp}
We plot for the following series of experiments the averaged $F$-measure, recall $\rec_V$, and the rank (cf. Sec.~\ref{sec:QualityMeas}), against the parameter varied when generating the synthetic data  (see Sec.~\ref{sec:SynthGen}). Error bars have length $2\sigma$. For every experiment, we generate eight matrices: two for each combination of dimensions $(n,m)\in\{(500,1600),\allowbreak(1600,500),\allowbreak(800,1000),\allowbreak(1000,800)\}$.
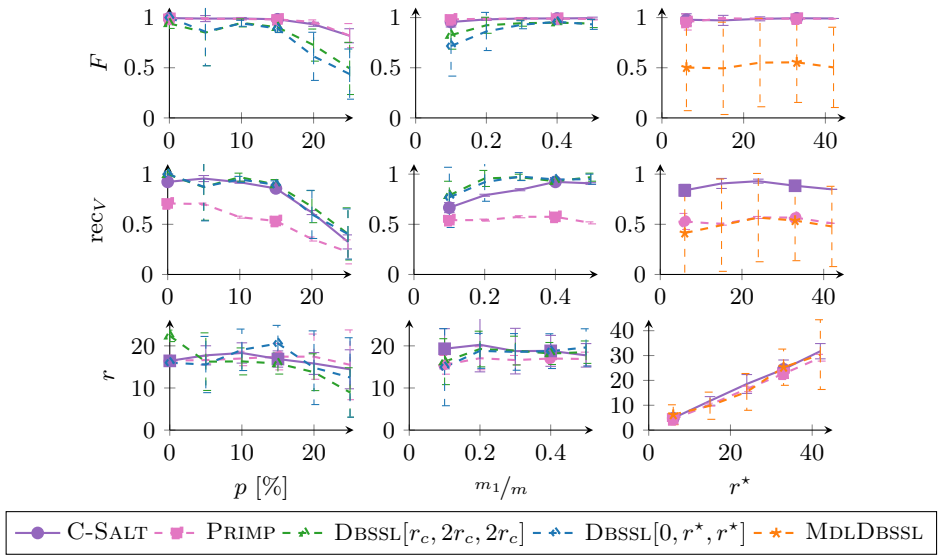
\begin{figure}[t!]
\centering
\pgfplotsset{
punkStyle/.style={cPunk,mark options ={cPunk},mark repeat={3}, thick, error bars/.cd,y dir = both, y explicit},
primpStyle/.style={cPrimp,dashed,mark options ={cPrimp},mark repeat={3}, thick, error bars/.cd,y dir = both, y explicit},
dbssl1Style/.style={cDBSSL1,dashed,mark options ={cDBSSL1},mark=triangle,mark repeat={3}, thick, error bars/.cd,y dir = both, y explicit},
dbssl2Style/.style={cDBSSL2,dashed,mark options ={cDBSSL2},mark=diamond,mark repeat={3}, thick, error bars/.cd,y dir = both, y explicit},
mdlDbsslStyle/.style={cMDLDBSSL,dashed,mark=star,mark options ={cMDLDBSSL},mark repeat={3}, thick, error bars/.cd,y dir = both, y explicit}
}
\begin{tikzpicture}[baseline]
		\begin{axis}[
        	axis lines = left,
            height=.25\linewidth,
            width=.33\linewidth,
			ylabel={$F$},
			xmin=0,xmax=25.5, ymin=0, ymax=1.1
			]
            \addplot+[punkStyle]  table[x=x,y=y,y error=std] {Punk_pF.dat};
            \addplot+[primpStyle]  table[x=x,y=y,y error=std] {Primp_pF.dat};
            \addplot+[dbssl1Style]  table[x=x,y=y,y error=std] {DBSSL1_pF.dat};
            \addplot+[dbssl2Style]  table[x=x,y=y,y error=std] {DBSSL2_pF.dat};
		\end{axis}
\end{tikzpicture}
\begin{tikzpicture}[baseline]
		\begin{axis}[
        	axis lines = left,
            height=.25\linewidth,
            width=.33\linewidth,
			xmin=0,xmax=0.52, ymin=0, ymax=1.1
			]
            \addplot+[punkStyle]  table[x=x,y=y,y error=std] {Punk_maF.dat};
            \addplot+[primpStyle]  table[x=x,y=y,y error=std] {Primp_maF.dat};
            \addplot+[dbssl1Style]  table[x=x,y=y,y error=std] {DBSSL1_maF.dat};
            \addplot+[dbssl2Style]  table[x=x,y=y,y error=std] {DBSSL2_maF.dat};
		\end{axis}
\end{tikzpicture}
\begin{tikzpicture}[baseline]
		\begin{axis}[
        	axis lines = left,
            height=.25\linewidth,
            width=.33\linewidth,
			xmin=0,xmax=45, ymin=0, ymax=1.1
			]
            \addplot+[punkStyle] table[x=x,y=y,y error=std]{Punk_rF.dat};
            \addplot+[primpStyle]  table[x=x,y=y,y error=std] {Primp_rF.dat};
            \addplot+[mdlDbsslStyle]  table[x=x,y=y,y error=std] {DBSSL_rF.dat};
		\end{axis}
\end{tikzpicture}
\begin{tikzpicture}[baseline]
		\begin{axis}[
        	axis lines = left,
            ylabel={$\rec_V$},
            height=.25\linewidth,
            width=.33\linewidth,
			xmin=0,xmax=25.5, ymin=0, ymax=1.1
			]
            \addplot+[punkStyle]  table[x=x,y=y,y error=std] {Punk_pRecV.dat};
            \addplot+[primpStyle]  table[x=x,y=y,y error=std] {Primp_pRecV.dat};
            \addplot+[dbssl1Style]  table[x=x,y=y,y error=std] {DBSSL1_pRecV.dat};
            \addplot+[dbssl2Style]  table[x=x,y=y,y error=std] {DBSSL2_pRecV.dat};
		\end{axis}
\end{tikzpicture}
\begin{tikzpicture}[baseline]
		\begin{axis}[
        	axis lines = left,
            height=.25\linewidth,
            width=.33\linewidth,
			xmin=0,xmax=0.52, ymin=0, ymax=1.1
			]
            \addplot+[punkStyle]  table[x=x,y=y,y error=std] {Punk_maRecV.dat};
            \addplot+[primpStyle]  table[x=x,y=y,y error=std] {Primp_maRecV.dat};
            \addplot+[dbssl1Style]  table[x=x,y=y,y error=std] {DBSSL1_maRecV.dat};
            \addplot+[dbssl2Style]  table[x=x,y=y,y error=std] {DBSSL2_maRecV.dat};
		\end{axis}
\end{tikzpicture}
\begin{tikzpicture}[baseline]
		\begin{axis}[
        	axis lines = left,
            height=.25\linewidth,
            width=.33\linewidth,
			xmin=0, xmax=45, ymin=0, ymax=1.1
			]
            \addplot+[primpStyle]  table[x=x,y=y,y error=std] {Primp_rRecV.dat};
            \addplot+[punkStyle]  table[x=x,y=y,y error=std] {Punk_rRecV.dat};
            \addplot+[mdlDbsslStyle]  table[x=x,y=y,y error=std] {DBSSL_rRecV.dat};
		\end{axis}
\end{tikzpicture}
\begin{tikzpicture}[baseline]
		\begin{axis}[
        	axis lines = left,
            height=.25\linewidth,
            width=.33\linewidth,
			xlabel={$p\ [\%]$},
			ylabel={$r$},
			xmax=25.5, xmin=0, ymin=0, ymax=26.2
			]
            \addplot+[primpStyle]  table[x=x,y=y,y error=std] {Primp_pR.dat};
            \addplot+[punkStyle] table[x=x,y=y,y error=std] {Punk_pR.dat};
            \addplot+[dbssl1Style]  table[x=x,y=y,y error=std] {DBSSL1_pR.dat};
            \addplot+[dbssl2Style]  table[x=x,y=y,y error=std] {DBSSL2_pR.dat};
		\end{axis}
\end{tikzpicture}
\begin{tikzpicture}[baseline]
		\begin{axis}[
        	axis lines = left,
            height=.25\linewidth,
            width=.33\linewidth,
			xlabel={$\nicefrac{m_1}{m}$},
			xmax=0.52, xmin=0, ymin=0, ymax=26.2
			]
            \addplot+[primpStyle]  table[x=x,y=y,y error=std] {Primp_maR.dat};
            \addplot+[punkStyle] table[x=x,y=y,y error=std] {Punk_maR.dat};
            \addplot+[dbssl1Style]  table[x=x,y=y,y error=std] {DBSSL1_maR.dat};
            \addplot+[dbssl2Style]  table[x=x,y=y,y error=std] {DBSSL2_maR.dat};
		\end{axis}
\end{tikzpicture}
\begin{tikzpicture}[baseline]
		\begin{axis}[
        	axis lines = left,
            height=.25\linewidth,
            width=.33\linewidth,
			xlabel={$r^\star$},
			xmax=45, xmin=0, ymin=0,
            legend columns =-1,
            legend entries={\textsc{C-Salt},\textsc{Primp},\textsc{Dbssl[$r_c,2r_c,2r_c$]},\textsc{Dbssl[$0,r^\star,r^\star$]},\textsc{MdlDbssl}},
            legend to name=named
			]
            \addplot+[punkStyle]  table[x=x,y=y,y error=std] {Punk_rR.dat};
            \addplot+[primpStyle]  table[x=x,y=y,y error=std] {Primp_rR.dat};
            \addplot[dbssl1Style] coordinates {(-1,-1)};
            \addplot[dbssl2Style] coordinates {(-1,-1)};
            \addplot+[mdlDbsslStyle]  table[x=x,y=y,y error=std] {DBSSL_rR.dat};
		\end{axis}
\end{tikzpicture}\\

\pgfplotslegendfromname{named}
\caption{Variation of noise (left column), class distribution $\nicefrac{m_1}{m}$ (middle column) and the rank (right column). The $F$-measure, recall of the matrix $V$ (both the higher the better) and the class-wise estimated rank of the calculated factorization is plotted against the varied parameter. Best viewed in color.}
\label{fig:noise}
\end{figure}

Fig.~\ref{fig:noise} contrasts the results of \textsc{C-Salt}, \textsc{Primp} and \textsc{Dbssl} in the two-class setting. For \textsc{Dbssl}, we consider two instantiations if the rank $r^\star$ is fixed. Both correctly reflect the number of planted specific and common patterns, yet the one rates class-specific alterations as separate patterns and the other counts every pattern with its class-specific alteration as a class-specific pattern. In the experiments varying the rank, we employ the MDL-based selection of the rank proposed for \textsc{Dbssl}. The input candidate constellations of class-specific and common patterns are determined according to the number of planted patterns, i.e., candidate rank constellations are a combination of $r_0\in \nicefrac{r^\star}{3}\pm\{5,0\}$ and  $r_{1}=r_2\in \{\nicefrac{kr^\star}{3}\mid k\in \{1,2,4\}\}$.

Fig.~\ref{fig:noise} shows the performance measures of the competing algorithms when varying three parameters: noise $p$ (left column), ratio of transactions in each class $\nicefrac{m_1}{m}$ (middle column) and rank $r^\star$ (right column).
We observe an overall high $F$-measure of \textsc{C-Salt} and \textsc{Primp}.
Both \textsc{Dbssl} instantiations also obtain high $F$-values, but only at lower noise levels and if one class is not very dominant over the other.
\textsc{C-Salt} and \textsc{Primp} differ most notably in the discovery of class specific alterations measured by $\rec_V$. \textsc{C-Salt} shows a similar recall as \textsc{Dbssl} if the noise is varied but a lower recall if classes are imbalanced. The ranks of returned factorizations by all algorithms lie in a reasonable interval, considering that class-specific alterations can also be interpreted as unattached patterns. Hence, a class-wise averaged rank between 16 and 24 is legitimate. When varying the number of planted patterns, the MDL selection procedure of the rank also yields correct estimations for \textsc{Dbssl}. However, the $F$-measure and recall of $V^\star$ decrease to 0.5 if the rank is not set to the correct parameters for \textsc{Dbssl}.

\begin{figure}[!t]
\centering
\pgfplotsset{
punkStyle/.style={cPunk,mark options ={cPunk},mark repeat={3}, thick, error bars/.cd,y dir = both, y explicit},
primpStyle/.style={cPrimp,dashed,mark options ={cPrimp},mark repeat={3}, thick, error bars/.cd,y dir = both, y explicit},
dbssl1Style/.style={cDBSSL1,dashed,mark options ={cDBSSL1},mark=triangle,mark repeat={3}, thick, error bars/.cd,y dir = both, y explicit},
dbssl2Style/.style={cDBSSL2,dashed,mark options ={cDBSSL2},mark=diamond,mark repeat={3}, thick, error bars/.cd,y dir = both, y explicit},
mdlDbsslStyle/.style={cMDLDBSSL,dashed,mark=star,mark options ={cMDLDBSSL},mark repeat={3}, thick, error bars/.cd,y dir = both, y explicit}
}

\begin{tikzpicture}
    \begin{groupplot}[group style={group size= 2 by 3, vertical sep=0.6cm},
    	height=.2\textwidth,
    	width=.4\textwidth,
        axis lines = left]
        \nextgroupplot[ylabel={$F$},xmin=0,xmax=25.5, ymin=0, ymax=1.1,
        legend to name=zelda]
        	\addplot+[punkStyle]  table[x=x,y=y,y error=std] {Punk_pC3RecV.dat};\addlegendentry{\textsc{C-Salt}};
            \addplot+[primpStyle]  table[x=x,y=y,y error=std] {Primp_pC3RecV.dat};\addlegendentry{\textsc{Primp}};
                \coordinate (top) at (rel axis cs:0,1);
        \nextgroupplot[xmin=0,xmax=25.5, ymin=0, ymax=1.1]
                \addplot+[punkStyle]  table[x=x,y=y,y error=std] {Punk_pC4F.dat};
            	\addplot+[primpStyle]  table[x=x,y=y,y error=std] {Primp_pC4F.dat};
        \nextgroupplot[ylabel={$\rec_V$},
			xmin=0,xmax=25.5, ymin=0, ymax=1.1]
                \addplot+[punkStyle]  table[x=x,y=y,y error=std] {Punk_pC3RecV.dat};
            	\addplot+[primpStyle]  table[x=x,y=y,y error=std] {Primp_pC3RecV.dat};
        \nextgroupplot[xmin=0,xmax=25.5, ymin=0, ymax=1.1]
                \addplot+[punkStyle]  table[x=x,y=y,y error=std] {Punk_pC4RecV.dat};
            	\addplot+[primpStyle]  table[x=x,y=y,y error=std] {Primp_pC4RecV.dat};
        \nextgroupplot[xlabel={$p\ [\%]$},
			ylabel={$r$},
			xmax=25.5, xmin=0, ymin=0, ymax=26.2]
                \addplot+[primpStyle]  table[x=x,y=y,y error=std] {Primp_pC3R.dat};
            	\addplot+[punkStyle] table[x=x,y=y,y error=std] {Punk_pC3R.dat};
        \nextgroupplot[xlabel={$p\ [\%]$},
			xmax=25.5, xmin=0, ymin=0,]
                \addplot+[punkStyle]  table[x=x,y=y,y error=std] {Punk_pC4R.dat};
            \addplot+[primpStyle]  table[x=x,y=y,y error=std] {Primp_pC4R.dat};
                \coordinate (bot) at (rel axis cs:1,0);
    \end{groupplot}
    \path (top)--(bot) coordinate[midway] (group center);
    \node[right=1em,inner sep=0pt] at(group center -| current bounding box.east) {\pgfplotslegendfromname{zelda}};
\end{tikzpicture}
\caption{Variation of noise for generated data matrices with three (left) and four classes (right). The $F$-measure, recall of the matrix $V$ (both the higher the better) and the class-wise estimated rank of the calculated factorization (between 16 and 24 can be considered correct) is plotted against the varied parameter. Best viewed in color.}
\label{fig:synthClass}
\end{figure}
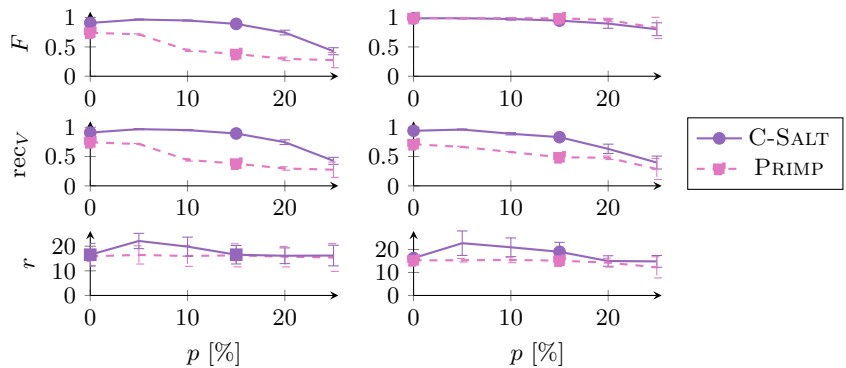
Fig.~\ref{fig:synthClass} displays the results of \textsc{Primp} and \textsc{C-Salt} when varying the noise for generated class-common and class-specific factorizations for three and four classes. The plots are similar to Fig.~\ref{fig:noise}. The more complex constellations of class-overarching outer products, which occur when more than two classes are involved, do not notably affect the ability to discover class-specific alterations by \textsc{C-Salt} and the planted factorization by \textsc{Primp} and \textsc{C-Salt}.
\subsection{Real-World Data Experiments}\label{sec:realWorld}
We explore the algorithms' behavior by three interpretable text-datasets depicted in Table~\ref{tbl:realWorld}. The datasets are composed by two classes to allow a comparison to \textsc{Dbssl}. The dimensions $m_1$ and $m_2$ describe how many documents belong to the first, respectively second class. Each document is represented by its occurring lemmatized words, excluding stop words. The dimension $n$ reflects the number of words which occur in 20 documents at least.
From the 20 Newsgroup corpus\footnote{\url{http://qwone.com/~jason/20Newsgroups/}}, we compose the \emph{Space-Rel} dataset by posts from\texttt{sci.space} and \texttt{talk.religion.misc}, and the \emph{Politics} dataset from \texttt{talk.politics.mideast} and \texttt{talk.politics.misc}. The \emph{Movie} dataset is prepared from a collection of 1000 negative and 1000 positive movie reviews\footnote{\url{http://www.cs.cornell.edu/People/pabo/movie-review-data/}}.

We consider two instantiations of \textsc{Dbssl}: \textsc{Dbssl1} is specified by $r_0=r_1=r_2=30$ and \textsc{Dbssl2} by $r_0=r_1=r_2=15$. For a fair comparison, we set a maximum rank of 30 for \textsc{C-Salt} and \textsc{Primp}. Therewith, the returned factorizations have a maximum rank of 90 for \textsc{Dbssl1}, 45 for \textsc{Dbssl2}, 30 for \textsc{Primp} and 60 for \textsc{C-Salt}. Note that \textsc{C-Salt} has the possibility to neglect $X$ and use mainly $V$ to reflect $cr=60$ class-specific outer products. In practice, we consider patterns $\Va_{\cdot s}{+}X_{\cdot s}$ as individual class-specific patterns if $|\Va_{\cdot s}|>|X_{\cdot  s}|$.
\begin{table}[!t]
	\centering
  \caption{Comparison of the amount of derived class-specific ($r_1,r_2$) and class-common patterns $(r_0)$, the overall rank $r=r_0+r_1+r_2$ and the RSS of the BMF (scaled by $10^4$) for real-world datasets. Values in parentheses correspond to factorizations where outer products with less than four items or transactions are discarded. The last two columns summarize characteristics of the datasets: number of rows belonging to the first and second class ($m_1$, $m_2$), number of columns ($n$) and density $d=|D|/(nm)$ in percent.}
  \label{tbl:realWorld}
  \resizebox{\columnwidth}{!}{%
	\begin{tabular}{crrrr>{\columncolor[gray]{0.95}}r>{\columncolor[gray]{0.95}}r>{\columncolor[gray]{0.95}}r>{\columncolor[gray]{0.95}}rrrrr}\toprule
    &\multicolumn{4}{c}{Space-Rel} &\multicolumn{4}{c}{\cellcolor[gray]{0.95}Politics}&\multicolumn{4}{c}{Movie}\\
    &\textsc{C-Salt}&\textsc{Primp}& \textsc{Dbssl1}&\textsc{Dbssl2}&\textsc{C-Salt}&\textsc{Primp}& \textsc{Dbssl1}&\textsc{Dbssl2}&\textsc{C-Salt}&\textsc{Primp}& \textsc{Dbssl1}&\textsc{Dbssl2}\\ \midrule
    $r$ & 29(28) & 30(30) & 40(7) & 18(6) & 41(40) & 30(30) & 57(20) & 42(15) & 26(25) & 30(27) & 27(4) & 12(4)\\
    $r_0$ & 4(3) & 8(8) & 19(1) & 7(1) & 10(10) & 8(8) & 16(2) & 5(0) & 25(25) & 29(27) & 21(1) & 6(0)\\
    $r_1$ & 9(9) & 8(8) & 13(4) & 8(4) & 19(18) & 15(15) & 27(14) & 18(11) & 1(0) & 1(0) & 3(1) & 3(1)\\
    $r_2$ & 16(16) & 14(14) & 8(2) & 3(1) & 12(12) & 7(7) & 14(4) & 19(4) & 0(0) & 0(0) & 3(2) & 3(3)\\
    RSS & 76(77) & 76(76) & 73(79) & 76(79) & 119(119) & 122(122) & 110(122) & 116(123) & 320(320) & 319(319) & 315(318) & 316(318)\\
    \bottomrule \midrule
    &$m_1$&$m_2$&$n$&$d[\%]$ &$m_1$&$m_2$&$n$&$d[\%]$ &$m_1$&$m_2$& $n$&$d[\%]$\\ 
    &622 & 980 & 2244 & 2.27 &936 & 775 & 2985 & 2.64 &998 & 997 & 4442 & 3.68\\ 
    \bottomrule
    \end{tabular}
    }
\end{table}

Table~\ref{tbl:realWorld} shows the number of class-specific and common patterns, and the resulting RSS. Since outer products involving only a few items or transactions either provide little insight or are difficult to interpret, we also state in parentheses the values concerning \textit{truncated factorizations}, i.e., outer products reflecting less than four items or transactions are discarded (glossing over the truncating of singletons, which is performed in both cases).

The untruncated factorizations obtained from \textsc{Dbssl} generally obtain a low RSS. However, when we move to the more interesting truncated factorizations, \textsc{Dbssl} suffers (the rank shrinks to less than a third for factorizations of \textsc{Dbssl2}). On the 20 News datasets this leads to a substantial RSS increase; \textsc{C-Salt} and \textsc{Primp} provide the lowest RSS in this case. We also observe, that the integration of the matrix $V$ by \textsc{C-Salt} empowers the derivation of more class-specific factorizations than \textsc{Primp}. Nevertheless, both algorithms describe the Movie dataset only by class-common patterns. We inspect these results more closely in the next section, showing that mining class-specific alterations points at exclusively derived class characteristics, especially for the Movie dataset.  
\subsection{Illustration of Factorizations}\label{sec:Interpret}
\begin{figure}[!t]
  \centering
  \begin{tabular}{ccc@{\hskip 0.1in}cc@{\hskip 0.1in}cc}
    & \multicolumn{2}{c}{\textsc{C-Salt}} & \multicolumn{2}{c}{\textsc{Primp}} & \multicolumn{2}{c}{\textsc{Dbssl}} \\
    \multirow{2}{*}{\rotatebox{90}{Space-Rel}  }
    & \includegraphics[width=0.14\columnwidth]{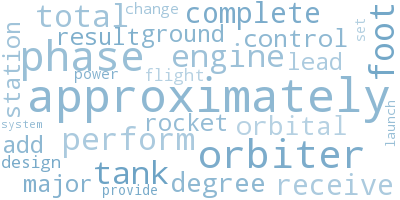}
    & \includegraphics[width=0.14\columnwidth]{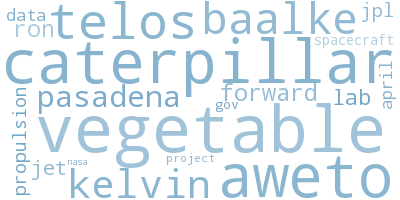}
    &  \includegraphics[width=0.14\columnwidth]{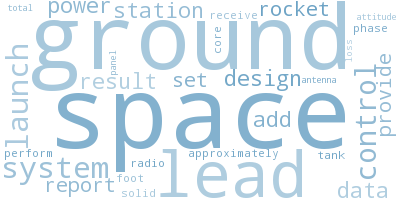}
    & \includegraphics[width=0.14\columnwidth]{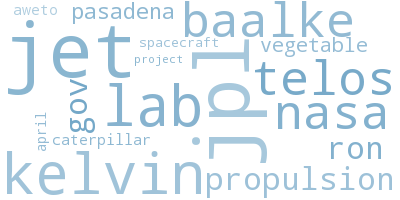}
    &  \includegraphics[width=0.14\columnwidth]{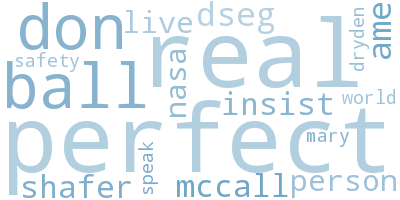}
    &  \includegraphics[width=0.14\columnwidth]{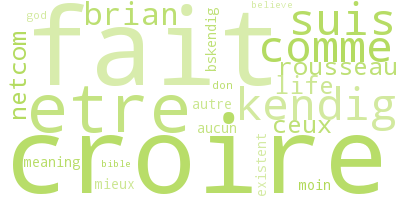}
    \\
    & \includegraphics[width=0.14\columnwidth]{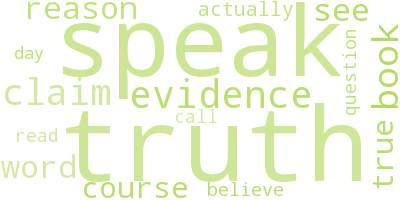}
    & \includegraphics[width=0.14\columnwidth]{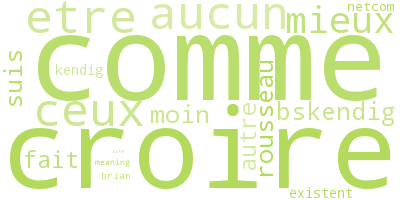}
    &  \includegraphics[width=0.14\columnwidth]{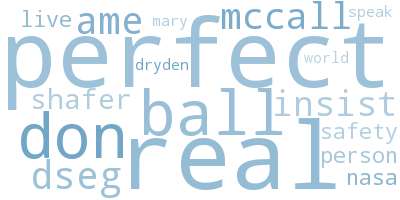}
    & \includegraphics[width=0.14\columnwidth]{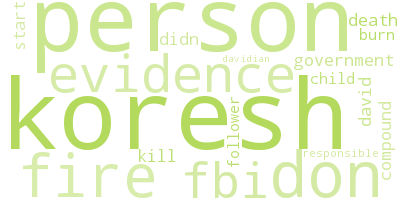}
    &  \includegraphics[width=0.14\columnwidth]{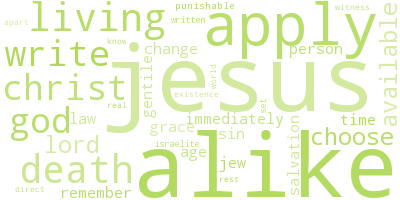}
    &  \includegraphics[width=0.14\columnwidth]{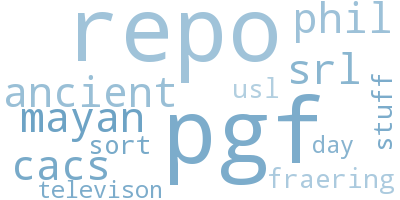}
     \\\\
     \multirow{2}{*}{\rotatebox{90}{Politics}  }
    & \includegraphics[width=0.14\columnwidth]{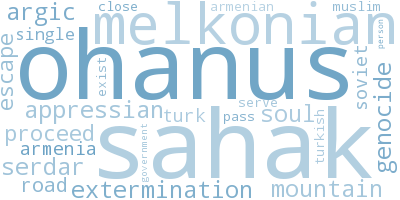}
    & \includegraphics[width=0.14\columnwidth]{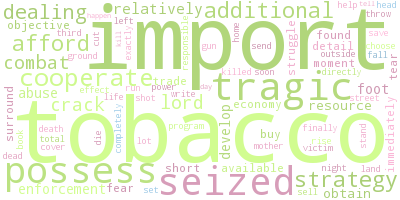}
    &  \includegraphics[width=0.14\columnwidth]{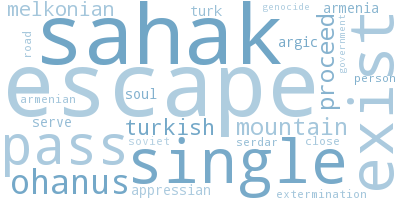}
    & \includegraphics[width=0.14\columnwidth]{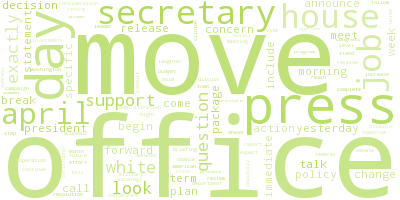}
    &  \includegraphics[width=0.14\columnwidth]{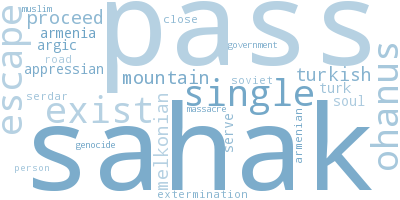}
    &  \includegraphics[width=0.14\columnwidth]{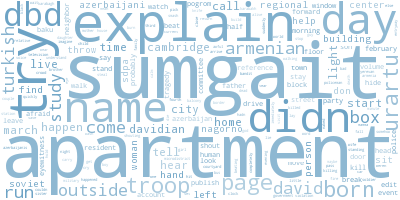}
    \\
    & \includegraphics[width=0.14\columnwidth]{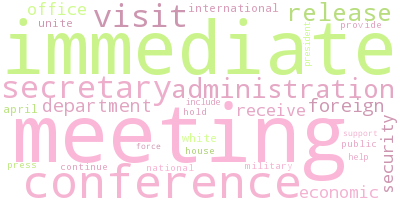}
    & \includegraphics[width=0.14\columnwidth]{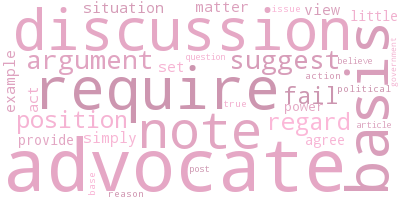}
    &  \includegraphics[width=0.14\columnwidth]{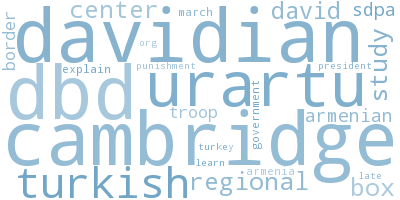}
    & \includegraphics[width=0.14\columnwidth]{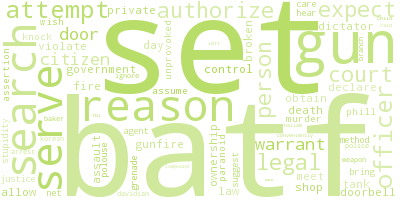}
    &  \includegraphics[width=0.14\columnwidth]{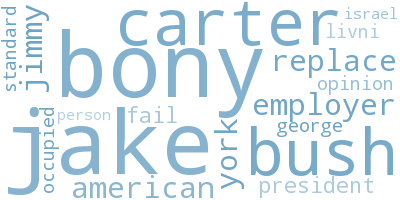}
    &  \includegraphics[width=0.14\columnwidth]{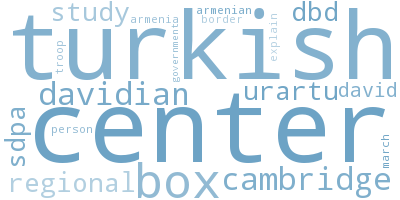}
     \\\\
    \multirow{2}{*}{\rotatebox{90}{Movie}  }
    & \includegraphics[width=0.14\columnwidth]{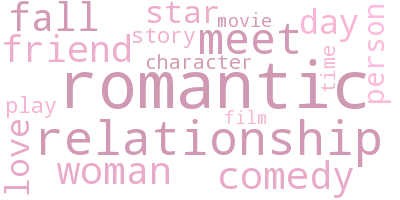}
    & \includegraphics[width=0.14\columnwidth]{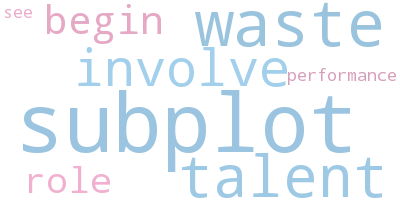}
    &  \includegraphics[width=0.14\columnwidth]{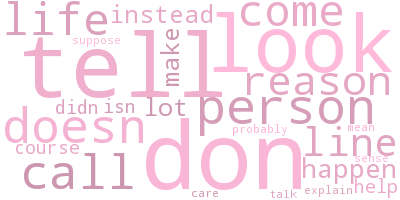}
    & \includegraphics[width=0.14\columnwidth]{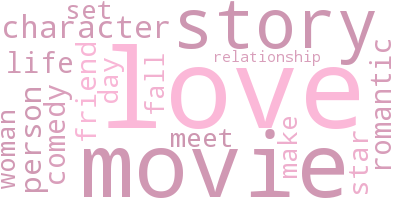}
    &  \includegraphics[width=0.14\columnwidth]{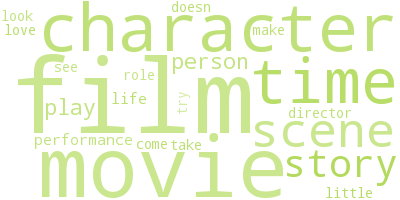}
    &  \includegraphics[width=0.14\columnwidth]{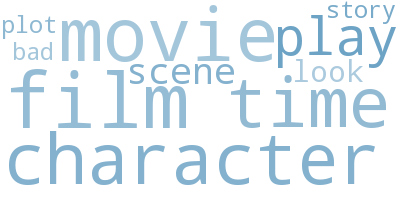}
    \\
    & \includegraphics[width=0.14\columnwidth]{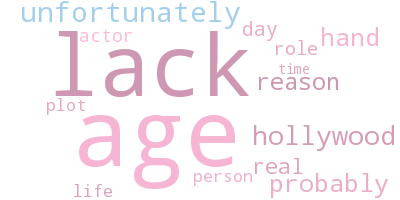}
    & \includegraphics[width=0.14\columnwidth]{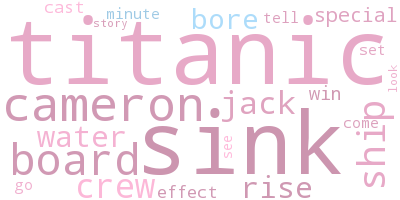}
    &  \includegraphics[width=0.14\columnwidth]{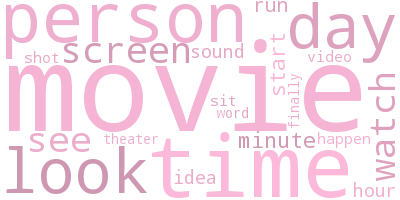}
    & \includegraphics[width=0.14\columnwidth]{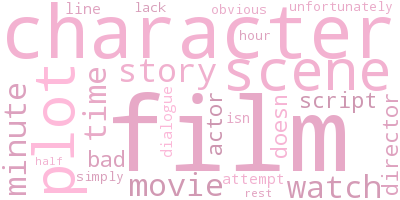}
    &  \includegraphics[width=0.14\columnwidth]{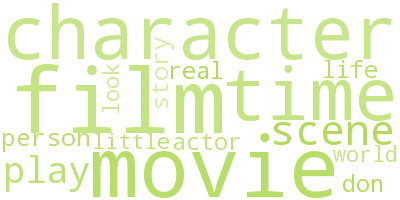}
    &  \includegraphics[width=0.14\columnwidth]{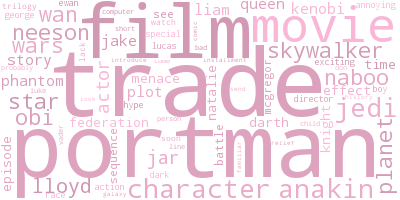}
  \end{tabular}
  \caption{Illustration of a selection of derived topics for the 20 News and Movie datasets. The size of a word reflects its frequency in the topic ($\sim Y_{\cdot s}^TD_{\cdot i}$) and the color its class affiliation: pink words are class-common, blue words belong to the first and green words to the second class. Best viewed in color.\label{fig:topics}}
\end{figure}
Let us inspect the derived most prevalent topics in the form of word clouds.
Fig.~\ref{fig:topics} displays for every algorithm the top four topics, whose outer product spans the largest area. Class-common patterns are colored pink whereas class-specific patterns are blue or green. Class-specific alterations within topics become apparent by differently colored words in one word cloud. We observe that the topics displayed for the 20-News data are mostly attributed to one of the classes. The topics are generally interpretable and even comparable among the algorithms (cf.\@ the first topic in the Politics dataset). Here, class-specific alterations of \textsc{C-Salt} point at the context in which a topic is discussed, e.g., the press release from the white house after a conference or meeting took place, whereby the latter may be discussed in both threads (cf.\@ the third topic for the Politics dataset).

The most remarkable contribution of class-specific alterations is given for the movie dataset. Generally, movie reviews addressing a particular genre, actors, etc., are not exclusively bad or good. \textsc{Primp} and \textsc{C-Salt} derive accordingly only common patterns. Here, \textsc{C-Salt} can derive the decisive hint which additional words indicate the class membership. We recall from Table~\ref{tbl:realWorld} that \textsc{DBSSL} returns in total four truncated topics for the Movie dataset. Thus, the displayed topics for the Movie dataset represent all the information we obtain from \textsc{DBSSL}. In addition, the topics display a high overlap in words, which underlines the reasonability of our assumption that minor deviations of major and common patterns can denote the sole class-distinctions.  
\subsection{Genome Data Analysis}\label{sec:gene}
The results depicted in the previous section are qualitatively easy to assess. We easily identify overlapping words and filter the important class characteristics from the topics at hand. 
In this experiment, the importance or meaning of features is unclear and researchers benefit from any summarizing information which is provided by the method, e.g., the common and class-specific parts of a pattern.
We regard the dataset introduced in~\cite{sangNature} representing the genomic profile of 18 Neuroblastoma patients. For each patient, samples are taken from three classes: \emph{normal} (N), \emph{primary tumor} (T) and \emph{relapse tumor  cell} (R). The data denotes loci and alterations taking place with respect to a reference genome. Alterations denote nucleotide variations such as $A\rightarrow C$, insertions ($C\rightarrow AC$) and deletions ($AC\rightarrow A$). One sample from each of the classes N and T is given for every patient ($m_N=m_T=18$), one patient lacks one and another has three additional relapse samples ($m_R=20$), resulting in $m=56$ samples.   
We convert the alterations into binary features, each representing one alteration at one locus (position on a chromosome). The resulting matrix has $n\approx 3.7$ million columns.
\begin{figure}[!t]
\parbox[t]{.5\textwidth}{\null
  \centering
  \includegraphics[trim={5cm 12cm  4cm 12cm},clip,scale=0.18]{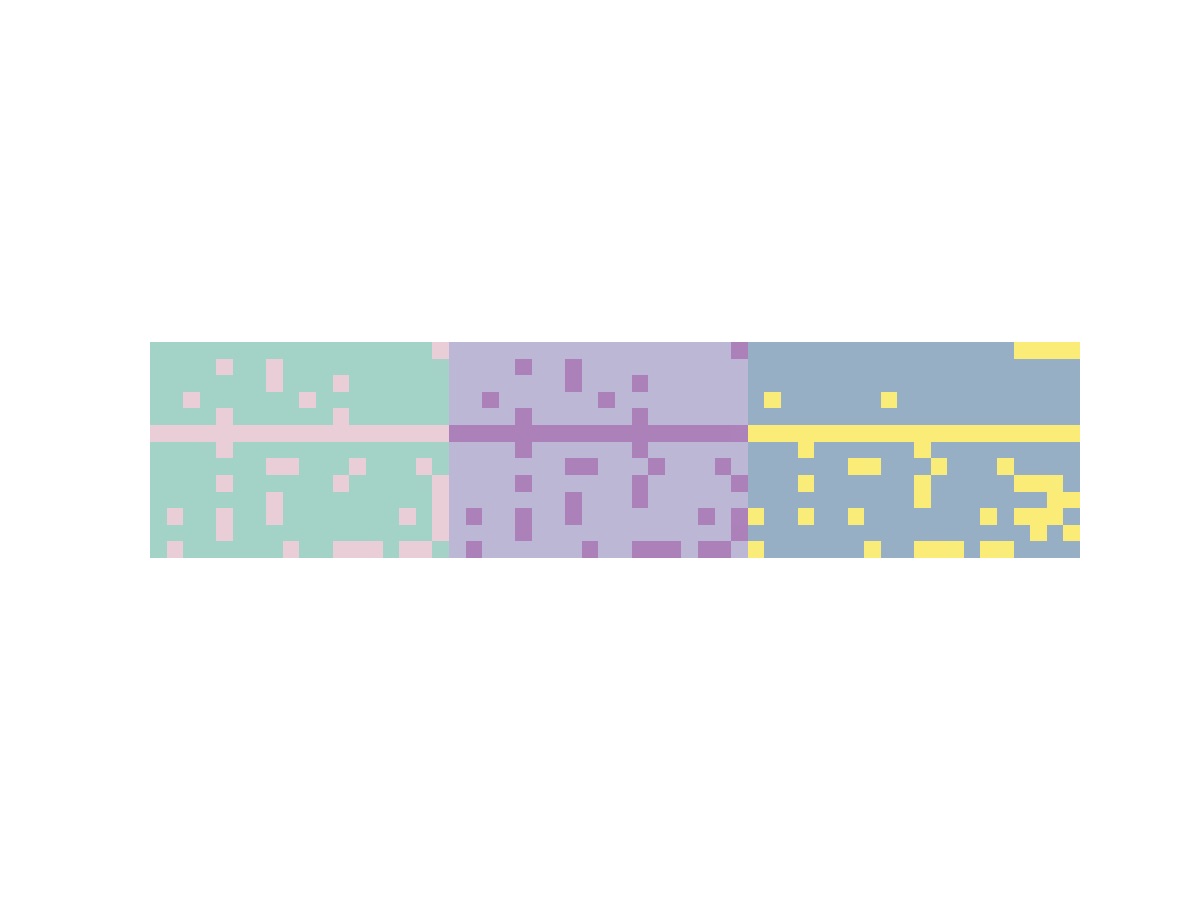}%
  \captionof{figure}{Transposed usage matrix returned by \textsc{C-Salt} on the genome dataset. Class-memberships are signalized by colors.}\label{fig:GenY}%
}\qquad
\parbox[t]{.46\textwidth}{\null
\centering
  \vskip-\abovecaptionskip
  \captionof{table}[t]{Average size and empirical standard deviation of patterns $(\cdot 10^3)$ and class-specific alterations $(\cdot 10^3)$.}\label{tbl:genePatterns}%
  \vskip\abovecaptionskip
\setlength{\tabcolsep}{7pt}
\resizebox{.46\textwidth}{!}{%
  \begin{tabular}{rrrr} \toprule
  $|X|$ & $|V^{(N)}|$ & $|V^{(T)}|$ &  $|V^{(R)}|$ \\\midrule
  $10.7\pm 96$ & $2.1\pm 2.5$ 
   & $3.6\pm 4.8$ 
   & $3.8\pm 6.6$ 
	\\\bottomrule
  \end{tabular}
}
}
\end{figure}

\textsc{C-Salt} returns on the genome data a factorization of rank 28, of which we omit sixteen patterns solely occurring in one patient. Fig.~\ref{fig:GenY} depicts the usage of the remaining twelve outer products, being almost identical for each class. Most notably, all derived patterns are class-common and describe the genetic background of patients instead of class characteristics. Table~\ref{tbl:genePatterns} summarizes the average length of patterns and corresponding class-specific alterations. We see that the average pattern reflects ten thousands of genomic alterations and that among the class-specific alterations, the ones which are attributed to relapse samples are highest in average. These results correspond to the evaluation in~\cite{sangNature}.

The information provided by \textsc{C-Salt} can not be extracted by existing methods. \textsc{Primp} yields only class-common patterns whose usage aligns with patients, regardless of classes. Running \textsc{Primp} separately on each class-related part $D^{(a)}$ yields factorizations of rank zero -- the genomic alignments between patients can not be differentiated from noise for such few samples.  
However, using the framework of \textsc{Primp} to minimize the RSS without any regularization, yields about 15 patterns for each part $D^{(a)}$. The separately mined patterns overlap over the classes in an intertwined fashion. The specific class characteristics are not easily perceived for such complex dependencies and would require further applications of algorithms which structure the information from the sets of vast amounts of features.
\section{Conclusion}
We propose \textsc{C-Salt}, an explorative method to simultaneously derive similarities and differences among sets of transactions, originating from diverse classes. \textsc{C-Salt} solves a Boolean Matrix Factorization (BMF) by means of numerical optimization, extending the method \textsc{Primp}~\cite{primp} to incorporate classes. We integrate a factor matrix reflecting class-specific alterations of outer products from a BMF (cf.\@ Definition~\ref{def:classSpec}). Therewith, we capture class characteristics, which are lost by unsupervised factorization methods such as \textsc{Primp}. Synthetic experiments show that a planted structure corresponding to our model assumption is filtered by \textsc{C-Salt} (cf.\@ Fig.~\ref{fig:noise}). Even in the case of more than two classes, \textsc{C-Salt} filters complex dependencies among them (cf.\@ Fig~\ref{fig:synthClass}). These experiments also show that the rank is correctly estimated. On interpretable text data, \textsc{C-Salt} derives meaningful factorizations which provide valuable insight into prevalent topics and their class specific characteristics (cf.\@ Table~\ref{tbl:realWorld} and Fig~\ref{fig:topics}). An analysis of genomic data underlines the usefulness of our new factorization method, yielding information which none if the existing algorithms can provide (cf.\@ Sec.~\ref{sec:gene}).

\paragraph{Acknowledgments}
Part of the work on this paper has been supported by Deutsche Forschungsgemeinschaft (DFG) within the Collaborative Research Center SFB 876 ``Providing Information by Resource-Constrained Analysis'', project C1
\url{http://sfb876.tu-dortmund.de}.

\appendix
\section[tbl]{Functions, Gradients and Lipschitz-Moduli}\label{app:func}
The functions, required by Algorithm~\ref{alg:C-Salt}, are stated in relation to $N=[N^{(a)}]_a$, as defined in Eq.~(\ref{eq:VBMF}).
$F$, stated in Sec.~\ref{sec:CSalt}, and its gradients are defined by
\begin{align*}
G(X,V,Y)&=-\sum_{s=1}^r(|Y_{\cdot s}|+1)\log\left(\frac{|Y_{\cdot s}|+1}{|Y|+r}\right) +|X^Tu|+ \sum_{a=1}^c|{V^{(a)}}^Tu| +|Y|,\\
\nabla_XF(X,V,Y)&=-\mu\sum_{a=1}^c{N^{(a)}}^T\Ya+u(0.5)^{1\times n},\\
\nabla_\Va F(X,V,Y)&=-\mu{N^{(a)}}^T\Ya+u(0.5)^{1\times n}+\nabla_\Va S(Y,V),\\
\nabla_\Ya F(X,V,Y)&=-\mu N^{(a)}X-\frac{1}{2}\left(\log\left(\frac{|Y_{\cdot s}|{+}1}{|Y|{+}r}\right)-1\right)_{js}+\nabla_{Y}^{(a)}S(Y,V),\\
  \nabla_\Va S(Y,V)&= D^TY +(1^{m_a\times n}-2\Da)^T\Ya\\
  \nabla_\Ya S(Y,V)&= \Da \left(\sum_{b\neq a}V^{(b)}\right)+(1^{m_a\times n}-\Da)\Va.
\end{align*}
The Lipschitz moduli are $M_{\nabla_X F}(Y,V)=\mu\|YY^T\|$, $M_{\nabla_\Va F}(X,Y)=\mu\|\Ya{\Ya}^T\|$ and $M_{\nabla_Y^{(a)} F}(X,V)=\mu\|(X+\Va)(X+\Va)^T\|+m_a$, $M_{\nabla_YF}(X,V)=\|(M_{\nabla\Va F}(X,Y))_a\|$.

\end{document}